\documentclass[10pt]{article} 

\usepackage[preprint]{tmlr}

\usepackage{multirow}
\usepackage{graphicx}
\usepackage{booktabs}
\usepackage[shortlabels,inline]{enumitem}
\usepackage{wrapfig}
\usepackage{microtype}


\usepackage{xcolor}
\usepackage{tcolorbox}
\usepackage{algorithm}
\usepackage{algpseudocode}
\usepackage{amsmath}
\usepackage{amssymb}

\definecolor{linkblue}{rgb}{0.21,0.49,0.74}
\usepackage[pagebackref,breaklinks,colorlinks,allcolors=linkblue]{hyperref}
\usepackage{url}

\def\month{MM}  
\def\year{YYYY} 
\def\openreview{\url{https://openreview.net/forum?id=XXXX}} 

\title{Optimize Your Sampling: Tuned Diffusion Sampling with Bayesian Optimization}

\author{\name Travis Zhang\thanks{Equal Contribution.} \email tz98@cornell.edu \\
        \addr Cornell University
        \AND
        \name Christian Belardi\footnotemark[1] \email ckb73@cornell.edu \\
        \addr Cornell University
        \AND
        \name Justin Lovelace \\
        \addr Cornell University
        \AND
        \name Jin Peng Zhou \\
        \addr Cornell University
        \AND
        \name Saebyeol Shin \\
        \addr Cornell University
        \AND
        \name Carla P. Gomes \\
        \addr Cornell University
        \AND
        \name Kilian Q. Weinberger \\
        \addr Cornell University}

\begin{document}
\maketitle

\begin{abstract}
Sampling from a diffusion model typically requires many forward passes through a large neural network, making generation computationally expensive. While much work has focused on efficient solvers and samplers, comparatively little attention has been paid to selecting the sampling timesteps themselves. A recent line of work optimizes theoretically derived surrogates for sample quality rather than the quality metric itself. We propose Optimizing Your Sampling (OYS), which instead treats timestep selection as a black-box optimization problem, optimizing the target metric directly with Bayesian optimization. OYS outperforms both the default schedules and those of Align Your Steps on text-to-image generation, and improves over the default schedules on inpainting and other image tasks, in both quantitative and human evaluations. OYS requires no additional training, is applicable even to distilled models, and improves both simple and sophisticated samplers such as Euler and DPM-Solver++. A 5-step OYS schedule retains 89\%--94\% of the quality of a 50-step schedule while reducing inference cost by 10x.
\end{abstract}

\section{Introduction}
\begin{figure*}[t!]
    \centering
    \includegraphics[width=\linewidth]{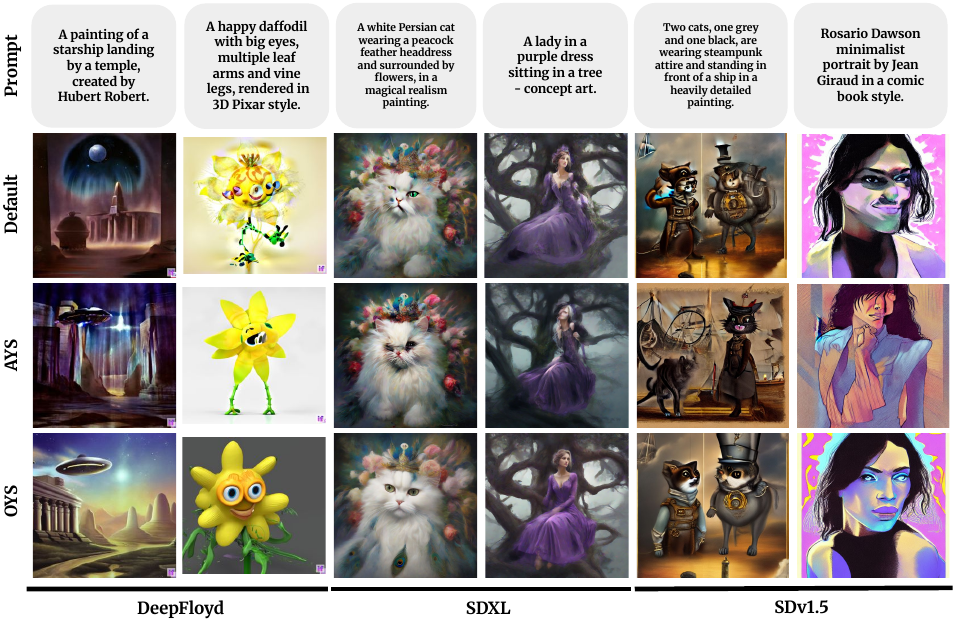}
    \caption{\textbf{Qualitative comparisons between Default, AYS \citep{sabour2024align}, and ours (OYS) using a 5 step sampling schedule.} All prompts are held-out test examples, unseen during schedule optimization for either method.}
    \label{figure:t2i_outputs}
\end{figure*}

Diffusion models \citep{ho2020denoising, song2021denoising, sohl2015deep} have achieved state-of-the-art performance across a wide range of generative tasks, including text-to-image synthesis \citep{ramesh2022hierarchical, rombach2022high, saharia2022photorealistic, peebles2023scalable}, super resolution \citep{Saharia2023SuperRes, saharia2022palette}, inpainting \citep{lugmayr2022repaint, corneanu2024latentpaint}, and video generation \citep{ho2022video, blattmann2023stable}. However, the iterative denoising process requires many forward passes through large neural networks, making generation computationally expensive.

Much of the effort to reduce this cost has focused on designing efficient samplers \citep{liu2022pseudo, lu2022dpm, lu2022dpmplusplus}, and developing few step models such as distilled models~\citep{salimans2022progressive, meng2023distillation, zhou2024simple, xie2024em, starodubcev2025scale, gu2023boot}, consistency models~\citep{song2023consistency, luo2023latent, lu2024simplifying}, and meanflow models~\citep{geng2026mean, geng2026improved}. Far less attention has been given to the choice of sampling timesteps, that is, which timesteps in the reverse diffusion process to use. Yet this choice has a substantial impact on quality, particularly at low step budgets where the default schedule degrades rapidly.

Prior work, Align Your Steps (AYS)~\citep{sabour2024align}, addresses this by formulating timestep selection as minimization of a KL divergence upper bound (KLUB) between the true generative process and its discretization. Fully minimizing this bound, however, need not minimize the mismatch between output distributions. Therefore AYS requires early stopping from a heuristic initialization~\citep{sabour2024align}. All the released AYS schedules, which were optimized at 10 steps, remain close to the default schedule in log-SNR space. We find that this is costly at low step budgets, where globally reallocating steps across the full timestep range yields larger gains than local refinement allows.

We propose Optimizing Your Sampling (OYS), which instead treats timestep selection as a black-box optimization problem and solves it with Bayesian optimization \citep{kushner1964new, zhilinskas1975single, mockus1975bayesian, jones1998efficient, osborne2009gaussian, garnett2023bayesian}. OYS searches the full configuration space without requiring gradient information, allowing the optimizer to reallocate steps across the entire timestep range rather than refining locally. It requires no additional training and applies to any diffusion model.

OYS consistently outperforms both AYS and the log-linearly downsampled default schedule on text-to-image generation, and the default schedule on inpainting and inverse image tasks, verified by both quantitative metrics and human evaluation; \autoref{figure:t2i_outputs} shows qualitative comparisons. Inspecting the optimized schedules reveals a consistent pattern: OYS allocates more steps to high-noise timesteps, in contrast to the approximately uniform log-SNR spacing of both AYS and the default schedule. OYS retains much of the quality of a 50-step schedule using only 5 steps. Further, OYS is efficient: for SDXL, tuning plateaus at approximately 17K five-step generations, 85K forward passes — over an order of magnitude fewer than the upper bound \citet{sabour2024align} report for AYS's tuning cost.
\begin{figure*}[t!]
    \centering
    \includegraphics[width=\linewidth]{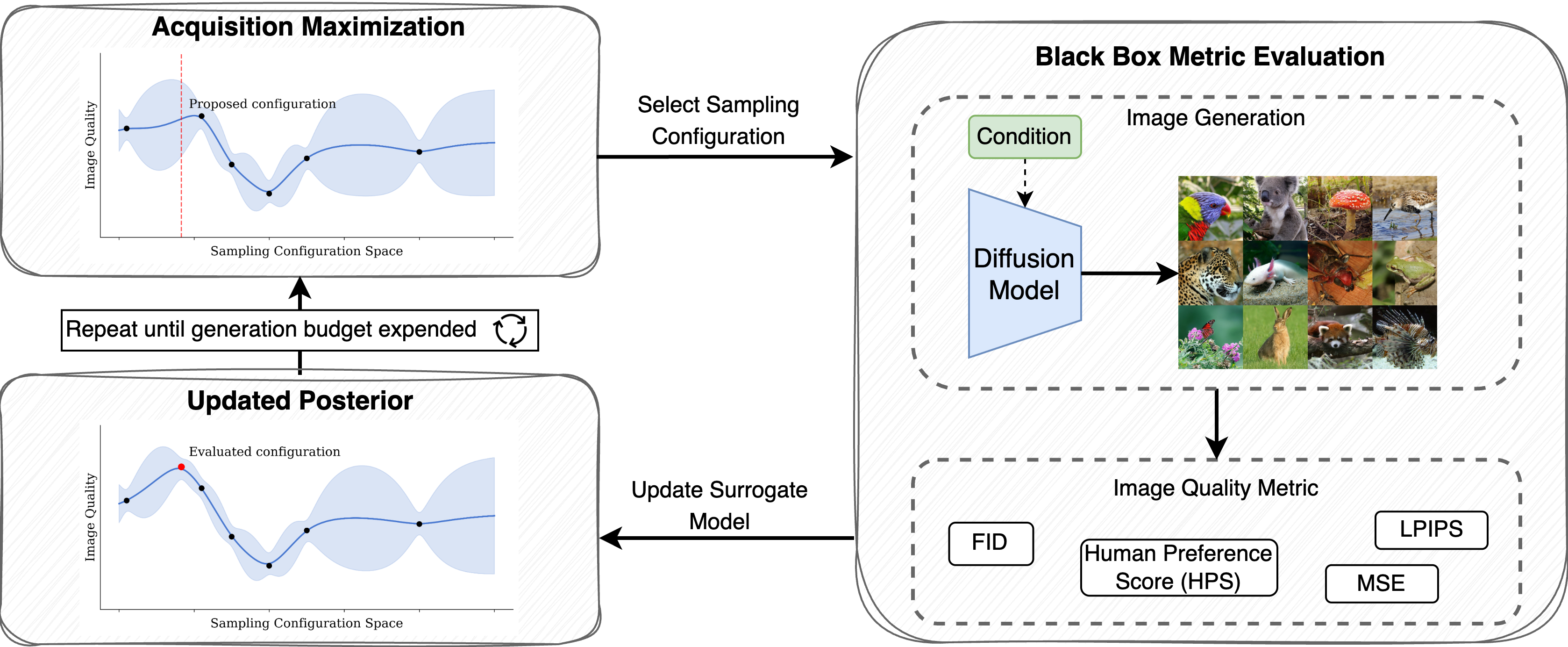}
    \caption{\textbf{Overview of our Optimize Your Sampling (OYS) framework using Bayesian Optimization.} The process begins with Acquisition Maximization (top left), where promising sampling configurations are identified. The selected configuration undergoes Black Box Metric Evaluation (right). This consists of generating images with the diffusion model and scoring them using metrics like FID and HPS. Results update the surrogate model's posterior distribution (bottom left), refining our understanding of the configuration space. This cycle repeats until convergence or budget exhaustion, efficiently discovering effective sampling parameters.}
    \label{fig:main_figure}
\end{figure*}

\section{Related Work}
\label{section:related_work}

Several prior works have explored customized sampling schedules or configurations. \citet{karras2022elucidating} present a unifying framework for diffusion model training and sampling that defines a number of parameters that control various aspects of both stochastic and deterministic sampling, including the schedule. The parameters $\sigma_{\min}$ and $\sigma_{\max}$ specify the bounds of the noise range, while $\rho$ controls how step sizes are distributed. However, finding the optimal configuration requires an expensive grid search, with each evaluation requiring 50,000 image generations.
\citet{karras2024analyzing} extend this framework with additional parameters including exponential moving average (EMA) length and guidance strength, again relying on grid search.

\citet{lugmayr2022repaint} were among the first to propose a custom sampling schedule, designing a denoising schedule for inpainting that repeatedly steps backward and forward in time — tracing a zigzag through the diffusion process — improving coherence between masked and unmasked regions at the cost of additional function evaluations.

\citet{watson2021learning} instead treat the schedule itself as an object of optimization, using dynamic programming to select the $K$-step subsequence that maximizes an evidence lower bound; the resulting schedules improve likelihood but not perceptual quality. \citet{watson2022learning} address this mismatch with Differentiable Diffusion Sampler Search (DDSS), which optimizes the degrees of freedom of a parametric family of samplers by backpropagating through the sampling process to maximize a perceptual score. This imposes two costs. First, the objective must admit gradients, so they optimize the Kernel Inception Distance as a differentiable stand-in for sample quality. Second, unrolling the sampling chain requires retaining its forward-pass state, whose memory grows linearly in the number of sampling steps and, as the authors note, quickly becomes infeasible for large denoiser architectures; DDSS must rematerialize the score function calls, recomputing them during the backward pass to trade this memory against additional compute. OYS shares the goal of optimizing perceptual quality directly, but treats the objective as a black box: each evaluation is an ordinary forward sampling pass, with no gradients flowing through the sampler. It therefore optimizes the target metric itself rather than a differentiable proxy, avoids both the memory of unrolling the sampler and the compute of rematerializing it, and accommodates non-differentiable operations such as the rounding of timesteps to integers for discrete-time models.

\citet{sabour2024align} propose Align Your Steps (AYS), which formulates schedule optimization as minimizing the Kullback-Leibler divergence Upper Bound (KLUB) between the true generative SDE and its discretization. 
The KLUB decomposes as a sum over the intervals between consecutive schedule points. AYS minimizes this total via a zeroth-order procedure that iteratively adjusts each intermediate timestep locally, with endpoints fixed and the schedule initialized from a heuristic schedule. They demonstrate that hand-crafted schedules such as~\citet{karras2022elucidating} are suboptimal, and their tuned schedules improve sample quality. However, because the KLUB is an upper bound rather than the output mismatch itself, minimizing it need not minimize that mismatch~\citep{sabour2024align}, and their procedure relies on early stopping to avoid over-optimizing it. Empirically, the resulting schedules remain close to the default schedule in log-SNR space (see~\autoref{figure:logsnr}). The released AYS schedules are 10-step; we adapt them to fewer steps via their prescribed log-linear interpolation.

Two further methods revisit timestep selection. Concurrent with our work, \citet{huang2026art} propose Adaptive Reparameterized Time (ART), an optimal control formulation that treats the speed of the sampler's clock as a control variable, redistributing computation along the trajectory so as to minimize the aggregate Euler discretization error, and solve the resulting problem with continuous-time reinforcement learning. In recent work, \citet{zhu2026hierarchical} propose the Hierarchical Schedule Optimizer (HSO), a bi-level method that alternates a global search over schedule initializations with local refinement of a midpoint error proxy, penalizing pathologically close timesteps to remain robust at very low step counts.

Both share the defining feature of AYS: they optimize a theoretically derived surrogate for sample quality — the KLUB for AYS, Euler discretization error for ART, and the midpoint error proxy for HSO — rather than the quantity of interest itself. Doing so makes their search inexpensive, as no images need be generated while optimizing, but it binds the resulting schedule to how faithfully the surrogate tracks perceptual quality. OYS sits at the opposite end of this trade-off. Because it evaluates the target metric directly, it incurs the cost of generating images during tuning, but it optimizes exactly the criterion the schedule is ultimately judged on, and applies unchanged to objectives for which no tractable surrogate exists — human preference scores, LPIPS, or task-specific reconstruction error. This cost is paid once per model and task and amortized over all subsequent sampling. Our approach also extends naturally beyond the timesteps themselves to the surrounding sampling parameters, such as guidance strength and EMA length, which surrogate-based objectives do not model.
\section{Optimize Your Sampling (OYS)}
\label{section:methods}

In this section, we describe the OYS framework. We present an overview of the optimization loop in~\autoref{fig:main_figure}.

At a high level, for a given task—such as text-to-image generation or inpainting—we assume access to a set of inputs $\mathcal{X}$ for tuning, a fixed pretrained diffusion model $m$, and a black-box evaluation metric $e$ that quantifies performance. OYS iteratively optimizes a set of sampling parameters $\mathbf{p} = \{p_1, \dots, p_M\}$ using Bayesian optimization, continuing until the optimization either converges or the budget is expended.

\paragraph{Optimizing Different Diffusion Formulations.} Diffusion models can be broadly categorized into two formulations: discrete-time, which defines the diffusion process as a fixed-length Markov chain of noise-addition steps, and continuous-time, which defines it as a differential equation over a continuous noise schedule. In both formulations, time parametrizes the noise level, and sampling requires selecting an ordered sequence of timesteps, the sampling schedule, at which to feed through the model. How this sampling schedule is represented varies: most models leave it unparameterized, specifying the timesteps directly, while some continuous-time models define it as a parametric family of functions with a few tunable parameters. We address each case below.

\paragraph{Direct schedule optimization.} Most diffusion models do not prescribe a parametric family for the sampling schedule. For these models, we optimize the schedule timesteps directly.

The key challenge is parameterizing the schedule for continuous optimization. While there is no strict requirement that sampling schedules be monotonically decreasing, standard practice suggests that the denoising trajectory should proceed from high to low noise. We enforce this with a cumulative product reparameterization: for each step $k$ in a $K$-step schedule, we optimize a continuous parameter $p_k \in [0, 1]$ and define the schedule as
\begin{align*}
    s(\mathbf{p}) &= \left[t_{\max}\prod_{i=1}^k p_i\right]_{k=1}^{K} \\
    &= \left[ t_{\max}p_1, (t_{\max}p_1p_2), \dots, (t_{\max}p_1p_2\dots p_K)\right]
\end{align*}

where $t_{\max}$ is the maximum timestep and $K\leq M$ to allow for additional hyperparameters beyond the schedule. Since each $p_k$ multiplies the running product by a value in $[0, 1]$, the schedule is monotonically decreasing by construction. Each parameter $p_k$ for $k>1$ is the ratio between consecutive timesteps, $t_k / t_{k-1}$.

For discrete-time models, whose timesteps must belong to a fixed set of integers (typically $\{0, \dots, 999\}$), we apply rounding, denoted by $q$, after scaling:
\begin{align*}
    s(\mathbf{p}) &= \left[q\Big(t_{\max}\prod_{i=1}^k p_i\Big)\right]_{k=1}^{K} \\
    &= \left[ q(t_{\max}p_1), q(t_{\max}p_1p_2), \dots, q(t_{\max}p_1p_2\dots p_K)\right].
\end{align*}

We note that this parameterization does not induce a uniform distribution over valid monotonic schedules. Because later timesteps are products of more $[0,1]$ terms, they concentrate near smaller values, biasing the search toward schedules that spend more steps at low noise levels. In practice, the Bayesian optimization surrogate learns to compensate for this non-uniformity, and we find that OYS consistently discovers schedules that allocate more steps to high-noise (large timestep) regions despite this bias, see \autoref{figure:logsnr}.

\paragraph{Parametric schedules.} The EDM family of diffusion models \citep{karras2022elucidating, karras2024analyzing} prescribes a functional form for their sampling schedules, parameterized by $\sigma_{\min}$, $\sigma_{\max}$, and $\rho$. \citet{karras2024analyzing} additionally tune the EMA length and classifier-free guidance strength. Since all parameters are real-valued and constrained to closed intervals, we treat each as an element in $\mathbf{p}$ and optimize them jointly. In this setting the timesteps themselves are tuned indirectly, through the parameters of the schedule's functional form.

\paragraph{Bayesian Optimization.} 
Our goal is to find the sampling configuration, $\mathbf{p}$, that minimizes the evaluation metric, $e$, on our tuning set, $\mathcal{X}$, for our model, $m$,
\begin{align*}
    \mathbf{p}^* &= \underset{\mathbf{p} \in \mathcal{P}}{\arg\min} \, e(m(\mathcal{X}; \mathbf{p})).
\end{align*}
For metrics where higher is better, such as HPS, we minimize the negated metric; we therefore phrase optimization as minimization throughout.

We use Bayesian optimization \citep{kushner1964new, zhilinskas1975single, mockus1975bayesian, jones1998efficient, osborne2009gaussian, garnett2023bayesian}, which sequentially selects and evaluates configurations using a Gaussian process to approximate the function $f(\mathbf{p}) = e(m(\mathcal{X}; \mathbf{p}))$.
The Gaussian process, fit to observed evaluations $\mathcal{D} = (\mathbf{p}^{(j)}, y^{(j)})_{j=1}^N$, provides a posterior distribution over the objective function, where $y^{(j)}$ is the observed value of sampling configuration $\mathbf{p}^{(j)}$. This posterior affords both mean and uncertainty estimates which are used to define an acquisition function that guides the search for promising sampling configurations. The initial configurations are chosen via Sobol sampling, with the number of Sobol samples set to twice the number of parameters. After evaluating the Sobol-sampled configurations, we fit a Gaussian process and switch to the qLogNEI~\citep{ament2023unexpected, letham2019constrained} acquisition function to select new candidate configurations.
\section{Experiments}
\label{section:experiments}

We benchmark sampling configurations found with OYS across a variety of image-domain tasks, comparing with relevant baselines when applicable. In all cases when we downsample a schedule, we use the log-linear interpolation procedure that \citet{sabour2024align} prescribe for adapting their schedules to other step counts. As the released AYS schedules are all 10-step schedules \citep{sabour2024align}, we obtain the few-step AYS baselines via their downsampling procedure. For all experiments, we evaluate on a held-out set of examples.

\paragraph{Evaluation Metrics.}
We evaluate with standard image quality metrics such as Fréchet Inception Distance (FID) \citep{heusel2017gans} and Peak Signal-to-Noise Ratio (PSNR) when applicable. However, these metrics do not capture alignment between text prompts and generated images.

To address this, \citet{wu2023humanv2} introduced the Human Preference Dataset v2 (HPD v2) and fine-tuned the ViT-L/14 version of a CLIP model on the dataset. Using this fine-tuned model, they define the Human Preference Score for a text-image pair $(T, I)$ as,
\begin{equation*}
    \begin{aligned}
        \text{HPS}(T,I) = \omega \frac{\mathcal{E}_{T}(T)}{||\mathcal{E}_{T}(T)||} \frac{\mathcal{E}_{I}(I)}{||\mathcal{E}_{I}(I)||}
    \end{aligned}
\end{equation*}
where $\mathcal{E}_{T}$ and $\mathcal{E}_{I}$ represent the CLIP text and image encoder respectively, and $\omega$ is a learned scaling parameter. Given a prompt $T$ and images $I_1$ and $I_2$ generated by two competing methods, the predicted preference win rate is computed from their HPS scores using the Bradley--Terry model,
\begin{equation*}
P(I_1 \succ I_2 \mid T) = \text{Softmax}\big(\text{HPS}(T,I_1), \text{HPS}(T,I_2)\big)_{I_1}.
\end{equation*}

We report empirical win rates from per-prompt comparisons: for each held-out prompt, we generate one image with each of the two schedules being compared, score both with HPS, and award the win to the higher-scoring image; the win rate is the fraction of prompts won. Unless otherwise noted, the comparison schedule is the default schedule at the same step count.

\paragraph{Text-to-Image.}
\begin{table}[t]
    \begin{minipage}[t]{0.48\textwidth}
        \centering
        \resizebox{\linewidth}{!}{%
        \begin{tabular}[t]{@{}ccccc@{}}
            \toprule
            \multirow{2}{*}{\textbf{Model}} & \multirow{2}{*}{\textbf{Method}} & \multirow{2}{*}{\textbf{Steps}} & \multirow{2}{*}{\textbf{\parbox{1.2cm}{\centering HPS\\Average$\uparrow$}}} & \multirow{2}{*}{\textbf{\parbox{2cm}{\centering HPS\\Win Rate$\uparrow$}}} \\
            & & & & \\
            \midrule
            \multirow{6}{*}{DeepFloyd}
             & Default & 50 & 0.267 & - \\
            \cmidrule(l){2-5}
             & Default & 10 & 0.258 & - \\
             & AYS & 10 & 0.261 & - \\
            \cmidrule(l){2-5}
             & Default & 5 & 0.215 & - \\
             & AYS & 5 & 0.225 & 65.39\% \\
             & OYS & 5 & 0.252 & 83.99\% \\
            \midrule
            \multirow{6}{*}{\parbox{1.5cm}{\centering Stable\\Diffusion\\v1.5}}
             & Default & 50 & 0.262 & - \\
            \cmidrule(l){2-5}
             & Default & 10 & 0.258 & - \\
             & AYS & 10 & 0.258 & - \\
            \cmidrule(l){2-5}
             & Default & 5 & 0.230 & - \\
             & AYS & 5 & 0.231 & 52.30\% \\
             & OYS & 5 & 0.238 & 59.29\% \\
            \midrule
            \multirow{6}{*}{\parbox{1.5cm}{\centering Stable\\Diffusion\\XL}}
             & Default & 50 & 0.275  & - \\
            \cmidrule(l){2-5}
             & Default & 10 & 0.266 & - \\
             & AYS & 10 & 0.264  & - \\
            \cmidrule(l){2-5}
             & Default & 5 & 0.229 & - \\
             & AYS & 5 & 0.210 & 25.29\% \\
             & OYS & 5 & 0.245 & 70.70\% \\
            \midrule
            \multirow{2}{*}{\parbox{1.5cm}{\centering SDXL\\Turbo}}
             & Default & 3 & 0.287 & - \\
             & OYS & 3 & 0.298 & 83.35\% \\
            \bottomrule
        \end{tabular}}
        \vspace{2pt}

        \caption{\textbf{Text-to-Image on COCO Captions.} Comparison of different models and methods on HPS Average and HPS Win Rate, evaluated on the COCO Captions validation split. Win Rates are computed with respect to the baseline scheduler at the same step count. DeepFloyd, Stable Diffusion v1.5, and Stable Diffusion XL use DPM-Solver++. For SDXL-Turbo, we benchmark the Euler Discrete (ED) sampler; Euler Ancestral (EA) results are included in the Appendix for completeness.}
        \label{table:text-to-image}
    \end{minipage}\hfill
    \begin{minipage}[t]{0.48\textwidth}
        \centering
        \resizebox{\linewidth}{!}{%
        \begin{tabular}[t]{@{}ccccc@{}}
            \toprule
            \multirow{2}{*}{\textbf{Model}} & \multirow{2}{*}{\textbf{Method}} & \multirow{2}{*}{\textbf{Steps}} & \multirow{2}{*}{\textbf{\parbox{1.2cm}{\centering HPS\\Average$\uparrow$}}} & \multirow{2}{*}{\textbf{\parbox{2cm}{\centering HPS\\Win Rate$\uparrow$}}} \\
            & & & & \\
            \midrule
            \multirow{2}{*}{\parbox{1.5cm}{\centering FLUX.1\\dev}}
             & Default & 5 & 0.251 & - \\
             & OYS & 5 & 0.283 & 82.69\% \\
            \midrule
            \multirow{2}{*}{\parbox{1.5cm}{\centering Qwen\\Image}}
             & Default & 5 & 0.174 & - \\
             & OYS & 5 & 0.193 & 77.19\% \\
            \bottomrule
        \end{tabular}}
        \vspace{2pt}

        \caption{\textbf{Text-to-Image on DiffusionDB.} Comparison of the default and OYS schedules on HPS Average and HPS Win Rate for FLUX.1-dev and QwenImage, tuned and evaluated on disjoint subsets of DiffusionDB. Both models use the Euler Discrete sampler. Win Rates are computed with respect to the default schedule at the same step count. Because these models are evaluated on a different prompt set, their HPS Averages are not directly comparable to those in \autoref{table:text-to-image}, and because both are run far below their default step counts, differences between the two models do not reflect their relative quality at full budgets.}
        \label{table:text-to-image-diffusiondb}

        \vspace{12pt}
        \resizebox{\linewidth}{!}{%
        \begin{tabular}[t]{@{}ccccc@{}}
            \toprule
            \multirow{2}{*}{\textbf{Model}} & \multirow{2}{*}{\textbf{Method}} & \multirow{2}{*}{\textbf{PSNR}$\uparrow$} & \multirow{2}{*}{\textbf{\parbox{1.2cm}{\centering HPS\\Average$\uparrow$}}} & \multirow{2}{*}{\textbf{\parbox{2cm}{\centering HPS\\Win Rate$\uparrow$}}} \\
            & & & & \\
            \midrule
            \multirow{2}{*}{\centering SDXL}
            & Default & 66.59 & 0.203 & - \\
            & OYS & 73.47 & 0.240 &  93.40\% \\
            \midrule
            \multirow{2}{*}{\centering SDv1.5}
            & Default & 64.46 & 0.213 & - \\
            & OYS & 67.34 & 0.237 & 85.10\% \\
            \bottomrule
        \end{tabular}}
        \vspace{2pt}

        \caption{\textbf{Inpainting.} PSNR, Average HPS, and HPS Win Rate on SDXL and SDv1.5 comparing Default and OYS schedules. HPS Win Rates are with respect to the Default schedule.}
        \label{table:inpainting}
    \end{minipage}
\end{table}

We evaluate OYS on six text-to-image models. On COCO Captions~\citep{chen2015microsoft}, we tune on the training split and evaluate on the validation split for DeepFloyd \citep{DeepFloydIF}, Stable Diffusion XL~\citep{podell2024sdxl}, and Stable Diffusion v1.5~\citep{rombach2022high} — the models with published AYS schedules \citep{sabour2024align}, our primary point of comparison — as well as SDXL-Turbo~\citep{sauer2025adversarial}, a model distilled for few-step sampling that tests whether schedule tuning complements distillation. On DiffusionDB~\citep{wang2022diffusiondb}, we tune FLUX.1-dev~\citep{labs2025flux1} and QwenImage~\citep{wu2025qwenimage} on one subset and evaluate on a held-out subset, comparing against their default schedules. We use Bayesian optimization to optimize the sampling timesteps to maximize HPS~\citep{wu2023humanv2}. For each round, we calculate HPS with a batch size of 128. Following \citet{sabour2024align}, we fix the sampler for DeepFloyd, Stable Diffusion XL, and Stable Diffusion v1.5 to DPM-Solver++~\citep{lu2022dpmplusplus}, while FLUX.1-dev, QwenImage, and SDXL-Turbo use the Euler Discrete sampler.

We report results on COCO Captions in \autoref{table:text-to-image} and on DiffusionDB in \autoref{table:text-to-image-diffusiondb}. As shown in \autoref{table:text-to-image}, at 10 steps schedule choice matters little: log-linearly downsampling the default schedule remains competitive with AYS across all three models. At 5 steps, however, performance degrades substantially and the schedules meaningfully separate. We therefore focus on this aggressive low-step regime, where schedule choice has the greatest leverage on quality. OYS discovers improved sampling schedules across all models, achieving higher win rates against the default and higher average HPS scores.

For SDXL-Turbo, the 3-step tuned schedule improves over the 3-step default schedule. For FLUX.1-dev and QwenImage, shown in \autoref{table:text-to-image-diffusiondb}, tuned 5-step schedules outperform default schedules, requiring only 5,376 and 2,560 generations respectively—equivalent in cost to just 960 and 256 generations at the models' respective 28- and 50-step default budgets. The two models' absolute HPS should therefore not be read as a model comparison: all values in \autoref{table:text-to-image-diffusiondb} reflect a 5-step budget, far below those defaults. These results span standard and distilled models, discrete-time and continuous-time formulations, and multiple samplers, suggesting that the benefits of schedule tuning with OYS are broadly applicable.

\begin{figure*}[t]
    \centering
    \includegraphics[width=\textwidth]{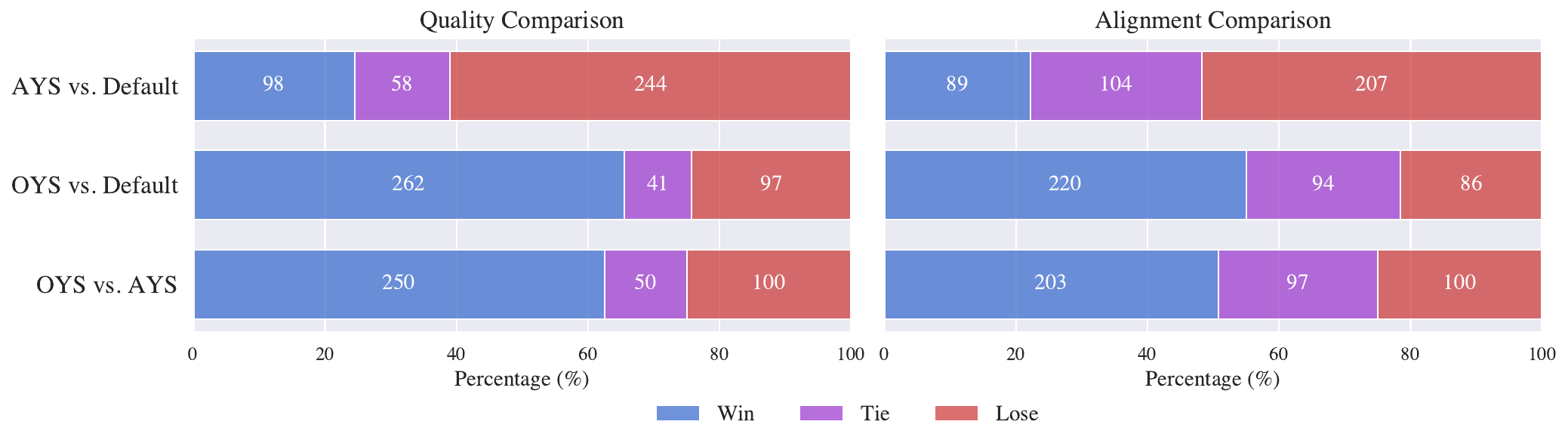}
    \caption{\textbf{Text-to-Image User Study.} Percent Win Rate on SDXL measuring image fidelity and image alignment. Significance was assessed with two-sided exact binomial tests on the win and lose counts, excluding ties, with all differences being highly significant ($p<0.001$). OYS significantly outperforms both AYS and default sampling in quality (68.8\% win rate vs. AYS, 70.6\% vs. Default; ties split equally) and in alignment (62.9\% win rate vs. AYS, 66.8\% vs. Default).}
    \label{figure:user_study}
\end{figure*}
To assess whether OYS's improvements correspond to human-perceived quality rather than being specific to the tuning metric, we conducted a human evaluation comparing generations from the 5-step schedules for AYS, OYS, and Default using SDXL. We crowdsourced responses from 58 participants assessing both image fidelity and alignment. Each participant received randomly shuffled pairs of images generated from either AYS, OYS, or Default, and was asked to choose whether image 1, image 2, or tie better answers the question. We provide an example of the rating interface in the Appendix. As shown in \autoref{figure:user_study}, OYS significantly outperforms both AYS and Default in image quality and alignment. Significance is assessed with a two-sided exact binomial test on the win and lose counts of each pairing, with ties excluded; every comparison is significant at $p<0.001$. For quality, OYS achieves a 68.8\% win rate against AYS and 70.6\% against Default, with ties split equally between methods. For alignment, OYS achieves a 62.9\% win rate against AYS and 66.8\% against Default. When AYS is log-linearly downsampled to 5 timesteps, it loses to the 5-step default schedule for SDXL. These human preference results are corroborated by performance on metrics that were never the tuning objective. Tuning on HPS improves FID substantially on DeepFloyd and SDXL. On SDv1.5, all 5-step schedules achieve comparable FID; we do not read much into differences this small, as zero-shot COCO FID only loosely tracks perceived quality \citep{hoogeboom2025simpler}. In no case does optimizing for human preference degrade distributional quality (see \autoref{table:text-to-image-fid} in the Appendix). As shown in the following sections, tuning on LPIPS for inpainting improves both PSNR and HPS (see \autoref{table:inpainting}), and tuning on MSE for the inverse image tasks improves HPS (see \autoref{table:prompt-diffusion}). Together, these results indicate that OYS's gains reflect genuine improvements in image quality.

\begin{figure*}[t]
    \centering
    \includegraphics[width=\linewidth]{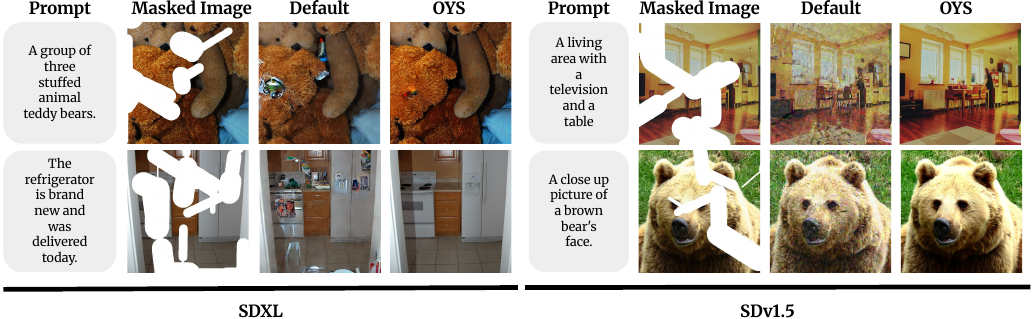}
    \caption{\textbf{Inpainting Results for SDv1.5 and SDXL against default and OYS schedule.} OYS reconstructions remain faithful to the unmasked regions, whereas the 5-step default schedule visibly corrupts the full image.}
    \label{figure:inpainting_outputs}
\end{figure*}

\paragraph{Inpainting.}
We evaluate OYS on inpainting. We optimize a 5-step sampling schedule for both Stable Diffusion 1.5 and Stable Diffusion XL on the training split of COCO Captions~\citep{chen2015microsoft} with a batch size of 64, evaluating on the validation split. For both models, we use the PNDMScheduler \citep{liu2022pseudo}. The optimization objective is LPIPS \citep{zhang2018unreasonable}, which measures perceptual similarity between the inpainted sample and the ground truth. The tuned schedules outperform the defaults on both models, improving PSNR by 6.9 dB on SDXL and 2.9 dB on SDv1.5 and winning 93.4\% and 85.1\% of HPS comparisons, respectively (see \autoref{table:inpainting} and \autoref{figure:inpainting_outputs}). Note that these models are trained to alter the full image, not just the masked region, enabling smoother transitions between inpainted and unmasked areas; under the 5-step default schedule this freedom becomes a failure mode — even unmasked regions are corrupted — while the OYS schedule preserves them.

\begin{figure*}[t]
    \centering
    \includegraphics[width=\textwidth]{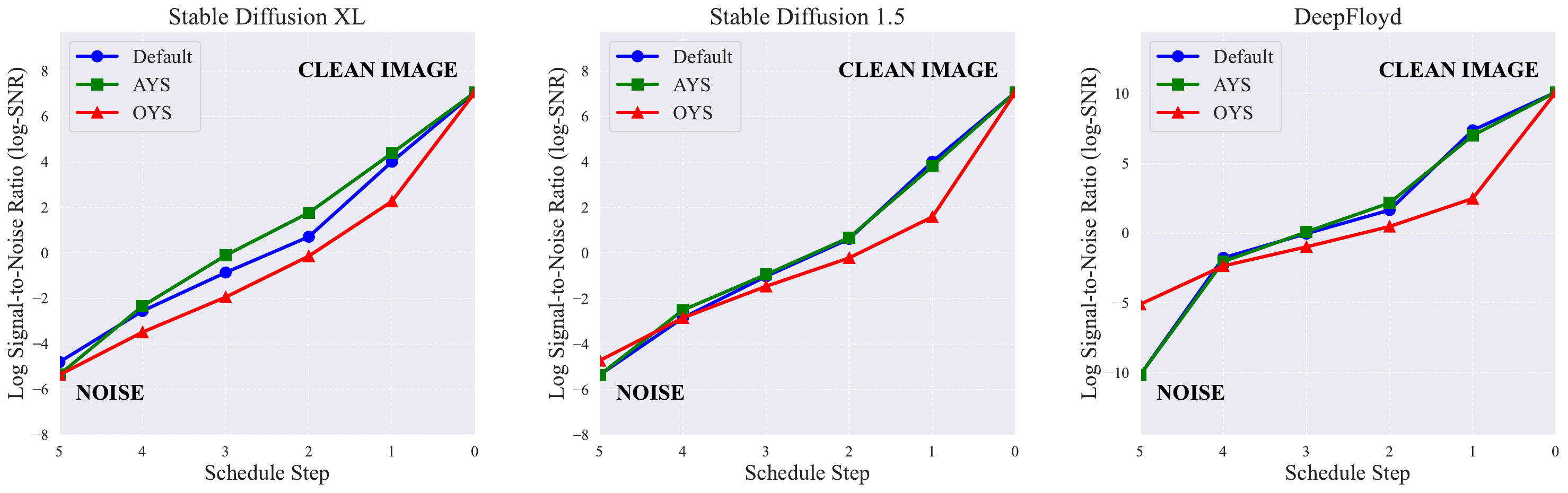}
    \caption{\textbf{5-step text-to-image sampling schedules in log-SNR space.} AYS closely tracks the default schedule's near-uniform log-SNR spacing, while OYS consistently places more steps at low log-SNR, i.e., high noise levels.}
    \label{figure:logsnr}
\end{figure*}

\paragraph{Schedule Analysis.} For any diffusion model, each timestep maps to a log Signal-to-Noise Ratio (log-SNR), providing a unified representation of noise levels across models \citep{kingma2023understanding}. We plot the schedules in log-SNR space in \autoref{figure:logsnr}. AYS and Default appear remarkably similar: AYS refines locally from a heuristic initialization with early stopping, which keeps its schedule close to the default. In contrast, OYS allocates more timesteps to higher noise levels, which improves downstream image quality and prompt alignment at low step budgets.

\begin{wraptable}[13]{r}{0.5\textwidth}
    \vspace{-\baselineskip}
    \centering
    \resizebox{\linewidth}{!}{%
    \begin{tabular}[t]{@{}ccccc@{}}
        \toprule
        \multirow{2}{*}{\textbf{Task}} & \multirow{2}{*}{\textbf{Method}} & \multirow{2}{*}{\textbf{PSNR}$\uparrow$} & \multirow{2}{*}{\textbf{\parbox{1.2cm}{\centering HPS\\Average$\uparrow$}}} & \multirow{2}{*}{\textbf{\parbox{2cm}{\centering HPS\\Win Rate$\uparrow$}}} \\
        & & & & \\
        \midrule
        \multirow{2}{*}{\parbox{1.5cm}{\centering Inverse\\HED}}
         & Default & 59.79 & 0.197 & - \\
         & OYS & 60.46 & 0.206 & 65.43\% \\
        \midrule
        \multirow{2}{*}{\parbox{1.5cm}{\centering Inverse\\Depth}}
         & Default & 58.51 & 0.198 & - \\
         & OYS & 59.05 & 0.202 & 57.62\% \\
         \midrule
        \multirow{2}{*}{\parbox{1.5cm}{\centering Inverse\\Segment}}
         & Default & 57.97 & 0.196 & - \\
         & OYS & 57.60 & 0.201 & 58.80\% \\
        \bottomrule
    \end{tabular}}
    \caption{\textbf{Prompt Diffusion.} PSNR, HPS, and Win Rate on Inverse HED, Inverse Depth, and Inverse Segmentation comparing Default and OYS schedules. HPS Win Rates are with respect to the Default Schedule.}
    \label{table:prompt-diffusion}
    \vspace{2mm}
\end{wraptable}

\begin{figure}[t]
    \begin{minipage}[t]{0.48\textwidth}
        \centering
        \includegraphics[width=\linewidth, trim=0 0 0 22, clip]{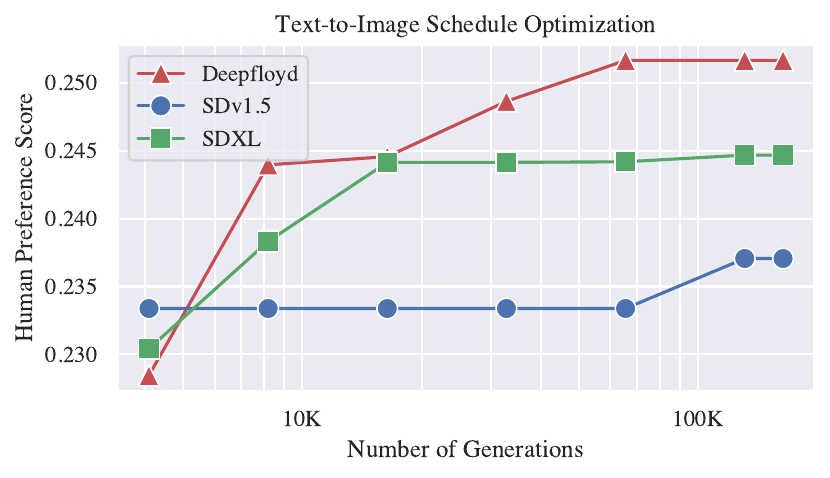}
        \caption{\textbf{Optimization efficiency for text-to-image schedule tuning across three diffusion models.} The graph shows Human Preference Score improvements versus number of generated images during optimization. All models achieve significant HPS gains with relatively few samples, typically plateauing after only 50-100K generations, highlighting OYS's efficiency relative to methods such as AYS, which can require substantially more generations.}
        \label{figure:efficiency_graph}
    \end{minipage}\hfill
    \begin{minipage}[t]{0.48\textwidth}
        \centering
        \includegraphics[width=\linewidth]{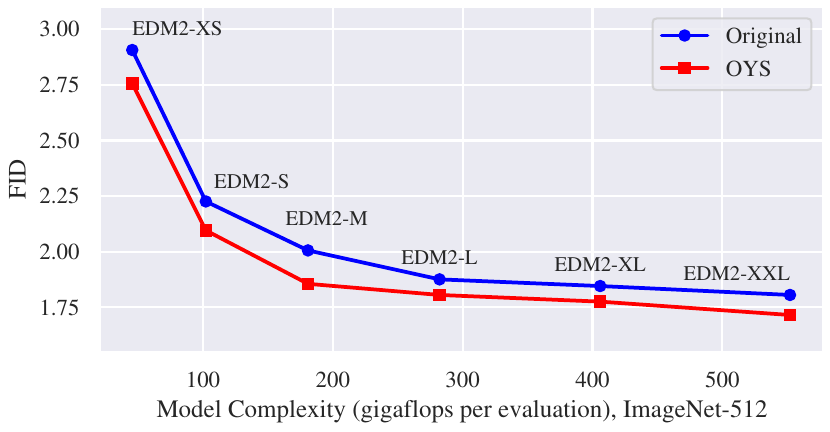}
        \caption{\textbf{ImageNet-512 Performance}. We compare between the original and OYS-optimized sampling configurations for the EDM2 model family. The graph shows FID scores (lower is better) against model complexity for ImageNet-512. Our OYS-optimized sampling configurations consistently outperform the original sampling configurations from \citet{karras2024analyzing} across all model sizes, from EDM2-XS to EDM2-XXL. }
        \label{figure:edm2_results}
    \end{minipage}
\end{figure}

\paragraph{Optimization Efficiency.} \autoref{figure:efficiency_graph} shows optimized schedule performance as a function of generations used during optimization. Performance for all models saturates before 170K images, with SDXL saturating before 17K generations. Since these are 5-step generations, the SDXL cost is equivalent to approximately 1.7K 50-step generations. In comparison, \citet{sabour2024align} report an upper bound on AYS's tuning cost—up to 300 iterations of 8,192 samples each, or approximately 2.4 million generations—one to two orders of magnitude more than OYS requires, depending on the model. Even discounting this bound by an order of magnitude, OYS remains cheaper for every model we tune.

\begin{figure*}[t]
    \centering
    \includegraphics[width=\textwidth]{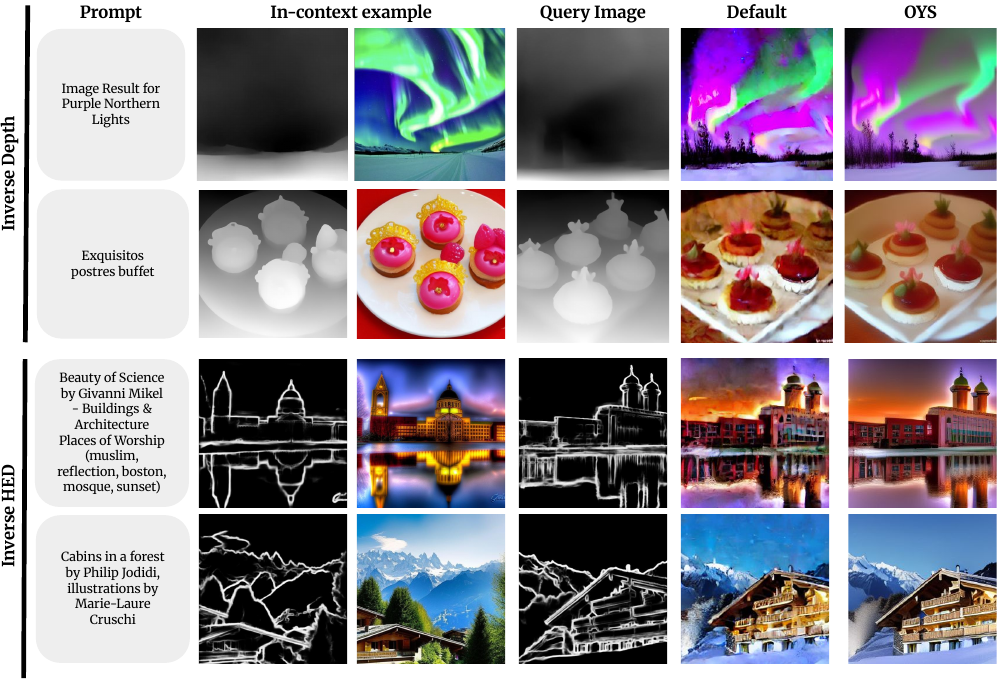}
    \caption{\textbf{Prompt Diffusion Results for Inverse Depth and Inverse HED against default and OYS schedule.} Best viewed in color.}
    \label{figure:prompt_diffusion_outputs}
\end{figure*}

\paragraph{Inverse HED, Inverse Depth, \& Inverse Segmentation.}
We explore sampling optimization for Prompt Diffusion \citep{wang2023context}, a model that performs multiple image generation tasks through in-context learning. It accepts both a text prompt, two images demonstrating the desired transformation, and a query image. We optimize schedules for three inverse image tasks: (1) Inverse HED transforms edge maps highlighting object boundaries into photorealistic images; (2) Inverse Depth converts depth maps into visually rich renderings; and (3) Inverse Segmentation transforms semantic segmentation maps into photorealistic images. Using the dataset from \citet{wang2023context}, we follow their setup to generate HED, depth, and segmentation maps. We optimize 5-step schedules on the validation split to minimize mean squared error between generated outputs and ground truth images, then evaluate on the test split. Consistent with the prior experiments, OYS achieves higher average HPS on all three tasks, with HPS win rates over the default schedule of 65.4\%, 57.6\%, and 58.8\% on Inverse HED, Inverse Depth, and Inverse Segmentation respectively, and improves PSNR on Inverse HED and Inverse Depth; on Inverse Segmentation PSNR is essentially unchanged (see \autoref{table:prompt-diffusion} and \autoref{figure:prompt_diffusion_outputs}).

\paragraph{EDM2 Image Generation.}
We investigate how much performance can be gained from tuning the sampling configurations for the EDM2 family of diffusion models \citep{karras2024analyzing}. \citet{karras2024analyzing} adopt the deterministic sampling configurations from \citet{karras2022elucidating}, found via a coarse-to-fine grid search. We optimize the parameters of their deterministic sampler discussed in~\autoref{section:related_work} — $\sigma_{\min}$, $\sigma_{\max}$, and $\rho$ — jointly with the EMA length and guidance strength, using the default guidance network provided by each model, to minimize FID estimated over 8,192 images per configuration.

OYS improves upon the grid-searched configurations of \citet{karras2024analyzing}, achieving FID improvements of 3.57\% to 7.00\% across model sizes (see \autoref{figure:edm2_results}) on ImageNet-512. The optimized configurations depart from the defaults in consistent ways: OYS narrows the sampled noise range at both ends, dropping $\sigma_{\max}$ from its default of 80 to roughly 20 for every model while raising $\sigma_{\min}$ five-fold for all but EDM2-XS, and it increases guidance for the smaller models but disables it entirely for EDM2-L and larger. We provide the full configurations and further analysis in the Appendix.

\section{Conclusion}
We present Optimize Your Sampling (OYS), a framework that casts diffusion sampling parameter selection as a black-box optimization problem, tuning the evaluation metric itself rather than a surrogate. It requires no retraining and applies to any pretrained model, sampler, or objective. OYS improves on both AYS and the default schedule for text-to-image generation, and on the default schedule for inpainting and inverse image tasks. The gains extend to SDXL-Turbo, a model already distilled for few-step sampling, showing that schedule tuning remains applicable even after distillation. A 5-step OYS schedule retains 89\%--94\% of the quality of a 50-step schedule at a tenth of the inference cost, and the improvements largely hold on metrics never used during tuning and are corroborated by human raters — evidence that they reflect genuine quality gains rather than exploitation of the tuning metric.

Inspecting the discovered configurations reveals consistent patterns that prevailing heuristics miss. In log-SNR space, short schedules benefit from allocating more steps to high-noise regions, in contrast to the approximately uniform spacing that both the default and AYS schedules exhibit. For the EDM2 family, where OYS tunes a parametric schedule rather than the timesteps themselves, the optimizer drives $\sigma_{\max}$ far below its default for every model size. Together these suggest that conventional spacing and noise-range heuristics leave real quality on the table at low step budgets, and that the parameters worth tuning extend beyond the timesteps alone.

OYS does not aim to produce universal schedules — optimal noise schedules are inherently domain-dependent \citep{chen2023importance} — but rather to make the already-necessary process of domain-specific tuning simpler and cheaper. Its cost is incurred once per model and task and amortized over all subsequent sampling. OYS thus offers a practical path to more efficient sampling across architectures and tasks.

\section*{Acknowledgements}
JL is supported by a Google PhD Fellowship. This work is also supported by the National Science Foundation through the AI Research Institutes program Award No. DMR-2433348; the National Institute of Food and Agriculture (USDA/NIFA); the Air Force Office of Scientific Research (AFOSR); New York-Presbyterian for the NYP-Cornell Cardiovascular AI Collaboration; and a Schmidt AI2050 Senior Fellowship.

{
    \small
    \bibliographystyle{tmlr}
    \bibliography{main}
}
\clearpage
\newpage
\appendix

\section{Additional Results}
We present detailed results for SDXL-Turbo in~\autoref{table:sdxl-turbo}. SDXL-Turbo is a distilled model designed for few-step inference, making it a challenging setting for schedule optimization as the default schedules are already highly compressed. We evaluate OYS with a 3-step Euler Discrete schedule against default schedules at both 1 and 3 steps using the Euler Discrete and Euler Ancestral samplers. OYS achieves the highest mean HPS of 0.298 and consistently outperforms all default baselines, with win rates ranging from 78.55\% to 83.94\%. Notably, OYS at 3 steps surpasses even the same-step-count defaults by a substantial margin, demonstrating that schedule optimization provides meaningful gains even for models specifically trained for few-step generation.

\begin{table}[!ht]
    \centering
    \begin{tabular}{@{}ccccc@{}}
        \toprule
        \textbf{Method} & \textbf{Sampler} & \textbf{Steps} & \textbf{\parbox{1.2cm}{\centering HPS\\Average$\uparrow$}} & \textbf{\parbox{3cm}{\centering OYS vs. Row\\HPS Win Rate$\uparrow$}} \\
        \midrule
        Default & EA/ED & 1 & 0.277 & 83.94\% \\
        Default & ED & 3 & 0.287 & 83.35\% \\
        Default & EA & 3 & 0.287 & 78.55\% \\
        OYS & ED & 3 & 0.298 & - \\
        \bottomrule
    \end{tabular}
    \caption{\textbf{SDXL-Turbo Results.} Mean HPS across the COCO Captions validation split and the HPS win rate of OYS (3-step, ED) against each default baseline. ED and EA denote Euler Discrete and Euler Ancestral, respectively.}
    \label{table:sdxl-turbo}
\end{table}

\citet{hoogeboom2025simpler} note that zero-shot FID on MSCOCO is known to not necessarily correspond to actual perceived performance. We nevertheless report FID for the text-to-image, inpainting, and Prompt Diffusion tasks. Text-to-Image and Inpainting were evaluated on the validation split of COCO Captions (see \autoref{table:text-to-image-fid} and \autoref{table:inpainting-fid}), and Prompt Diffusion on the test split of the dataset from \citet{wang2023context} (see \autoref{table:prompt-diffusion-fid}).

At 5 steps, OYS substantially improves FID over both the default and AYS schedules on DeepFloyd and SDXL. On SDv1.5, all three 5-step schedules lie within 0.7 FID of one another, with OYS slightly behind the baselines despite winning the HPS comparisons in \autoref{table:text-to-image} — consistent with the loose coupling between COCO FID and perceived quality noted above.

\begin{table}[!ht]
    \begin{minipage}[t]{0.48\textwidth}
    \centering
    \begin{tabular}[t]{@{}lccc@{}}
        \toprule
        \textbf{Model} & \textbf{Method} & \textbf{Steps} & \textbf{FID $\downarrow$} \\
        \midrule
        \multirow{6}{*}{DeepFloyd}
         & Default & 50 & 27.732 \\
        \cmidrule(l){2-4}
         & Default & 10 & 25.834 \\
         & AYS & 10 & 25.496 \\
        \cmidrule(l){2-4}
         & Default & 5 & 34.912 \\
         & AYS & 5 & 27.613 \\
         & OYS & 5 & 25.681 \\
        \midrule
        \multirow{6}{*}{\parbox{2cm}{\centering Stable\\Diffusion\\XL}}
         & Default & 50 & 25.182 \\
        \cmidrule(l){2-4}
         & Default & 10 & 26.130 \\
         & AYS & 10 & 24.584 \\
        \cmidrule(l){2-4}
         & Default & 5 & 31.264 \\
         & AYS & 5 & 36.757 \\
         & OYS & 5 & 27.088 \\
        \midrule
        \multirow{6}{*}{\parbox{2cm}{\centering Stable\\Diffusion\\v1.5}}
         & Default & 50 & 26.051 \\
        \cmidrule(l){2-4}
         & Default & 10 & 24.316 \\
         & AYS & 10 & 24.057 \\
        \cmidrule(l){2-4}
         & Default & 5 & 24.362 \\
         & AYS & 5 & 24.182 \\
         & OYS & 5 & 24.865 \\
        \bottomrule
    \end{tabular}
    \caption{\textbf{Text-to-Image FID.} FID on the COCO Captions validation split for DeepFloyd, SDXL, and SDv1.5 across schedules and step counts.}
    \label{table:text-to-image-fid}
    \end{minipage}\hfill
    \begin{minipage}[t]{0.48\textwidth}
    \centering
    \begin{tabular}[t]{@{}ccc@{}}
        \toprule
        \textbf{Model} & \textbf{Method} & \textbf{FID $\downarrow$} \\
        \midrule
        \multirow{2}{*}{\centering SDXL}
        & Default & 31.80 \\
        & OYS & 6.85 \\
        \midrule
        \multirow{2}{*}{\centering SDv1.5}
        & Default & 64.83 \\
        & OYS & 3.98 \\
        \bottomrule
    \end{tabular}
    \caption{\textbf{Inpainting FID.} FID on the COCO Captions validation split for SDXL and SDv1.5 with 5-step schedules.}
    \label{table:inpainting-fid}

    \vspace{12pt}
    \begin{tabular}[t]{@{}ccc@{}}
        \toprule
        \textbf{Task} & \textbf{Method} & \textbf{FID $\downarrow$} \\
        \midrule
        \multirow{2}{*}{Inverse HED}
         & Default & 40.93 \\
         & OYS & 9.05 \\
        \midrule
        \multirow{2}{*}{Inverse Depth}
         & Default & 34.50 \\
         & OYS & 12.89 \\
         \midrule
         \multirow{2}{*}{Inverse Segmentation}
         & Default & 32.42 \\
         & OYS & 13.57 \\
        \bottomrule
    \end{tabular}
    \caption{\textbf{Prompt Diffusion FID.} FID on the test split of the dataset from \citet{wang2023context} for the three inverse image tasks with 5-step schedules.}
    \label{table:prompt-diffusion-fid}
    \end{minipage}
\end{table}

\section{User Study}

\begin{figure*}[t]
\centering
\begin{tcolorbox}[title=\textbf{User Study Instructions}, 
                 colback=white, 
                 colframe=black, 
                 fonttitle=\bfseries\large,
                 width=0.95\textwidth,
                 toptitle=1mm,
                 bottomtitle=1mm,
                 top=1mm,
                 bottom=1mm]
\textbf{Basic Rules and Regulations}
We will provide you a pair of images, which consists of images generated from two different schedules. Please consider the prompts and images from the perspectives of universal and personal aesthetic appeal. This task mainly involves two aspects: text-image alignment and image quality. Although we encourage and value personal preference, it's important to consider the following fundamental principles when balancing the two aspects or facing a dilemma:
\begin{enumerate}[leftmargin=*, itemsep=0pt, parsep=0pt]
    \item For Question 1, we want to focus only on the image quality itself. The prompt here doesn't matter and we want you to solely compare the quality of the generations between the pairs of images.
    \item For Question 2, try to only focus on how aligned the prompt is to the image. Try to disregard the image quality, and focus on the semantic information involved.
    \begin{enumerate}[leftmargin=*, itemsep=0pt, parsep=0pt]
        \item If there is any content you are not familiar with, we recommend you to search for sample images and explanations online.
    \end{enumerate}
\end{enumerate}

\vspace{0.1em}
\noindent\rule{\textwidth}{0.4pt}

\begin{center}
\textbf{Example Image Pair}
\setlength{\tabcolsep}{8pt}
\begin{tabular}{cc}
\includegraphics[width=0.45\textwidth]{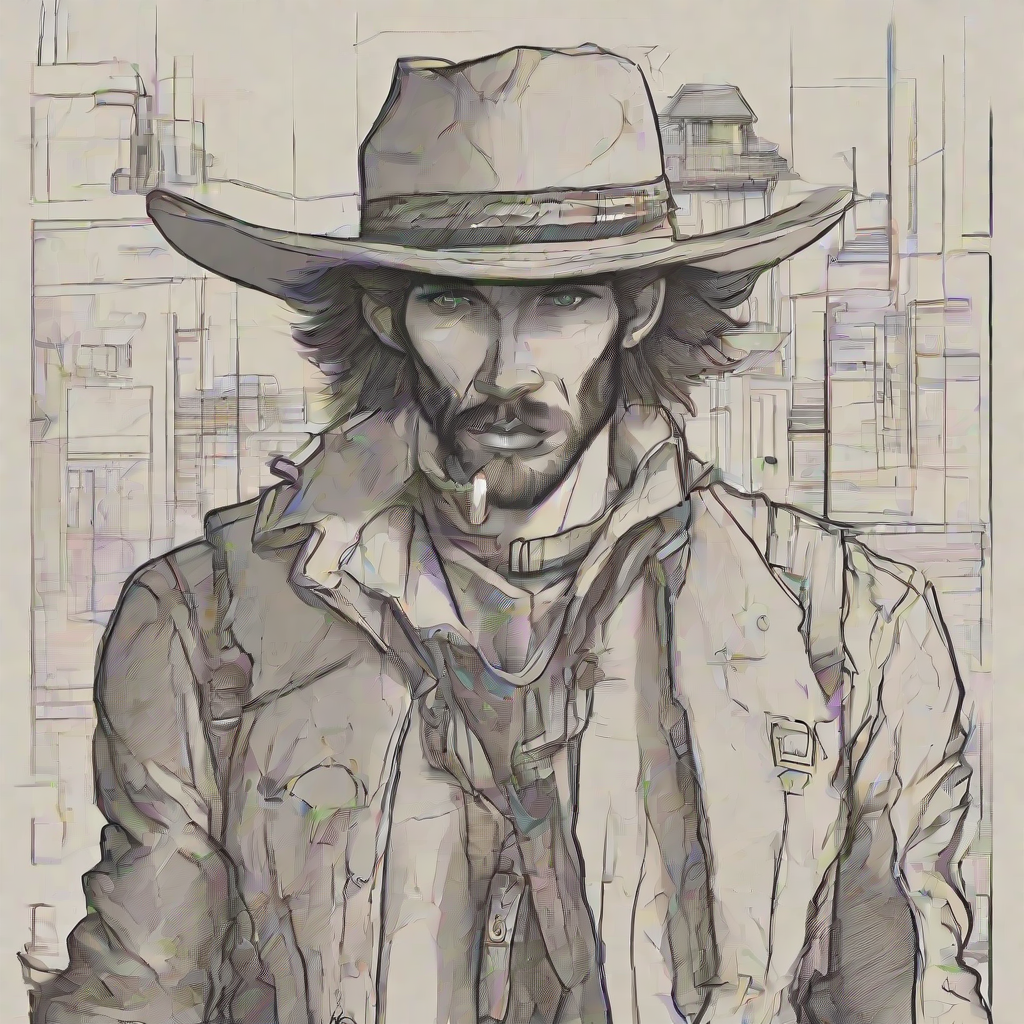} & 
\includegraphics[width=0.45\textwidth]{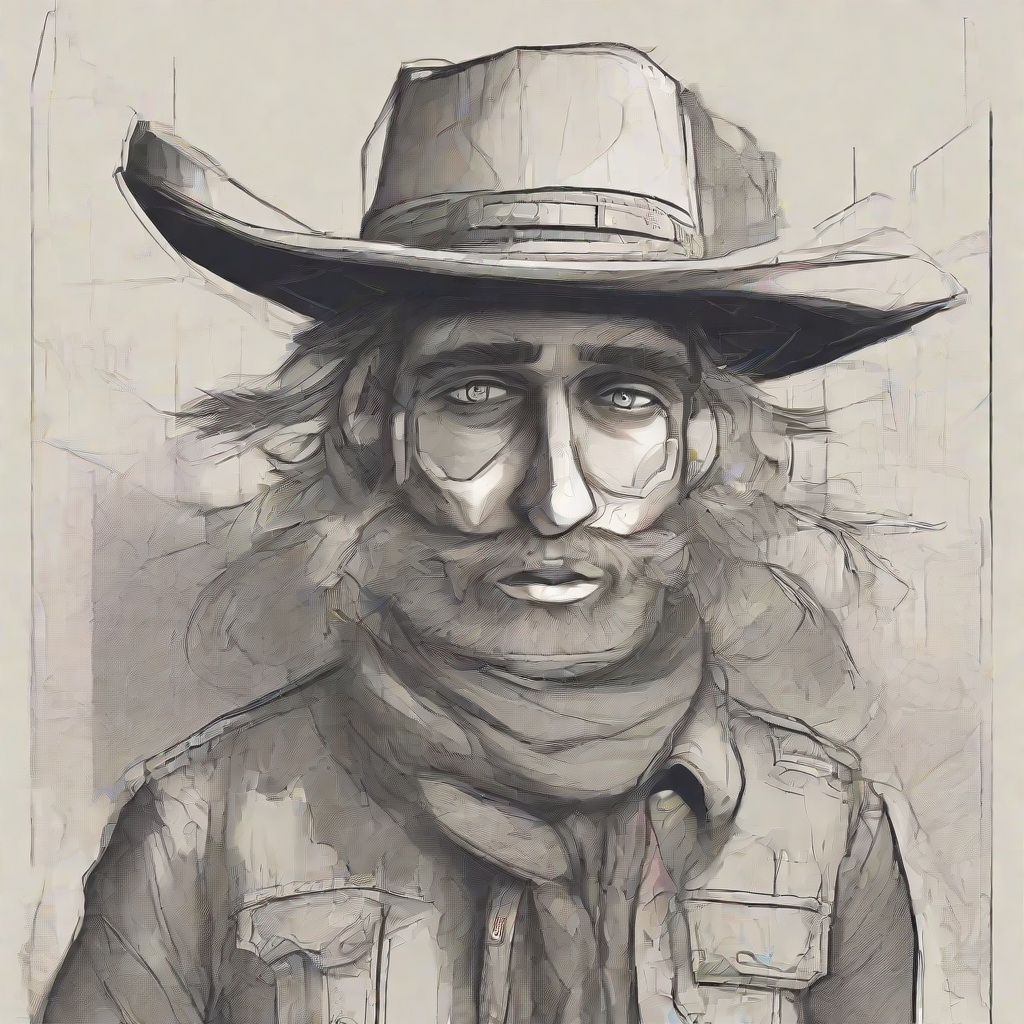} \\
Image 1 & Image 2
\end{tabular}
\end{center}

\textbf{Question 1:} Ignoring the prompt, which image is better in terms of quality?

\vspace{.4em}
\begin{tabular}{ll}
$\bigcirc$ & \textbf{Image 1} \\
$\bigcirc$ & \textbf{Image 2} \\
$\bigcirc$ & \textbf{Tie} (both are equally good/bad)
\end{tabular}

\vspace{0.2em}
\noindent\rule{\textwidth}{0.4pt}

\textbf{Question 2:} Which image better aligns with the prompt?
\begin{tcolorbox}[colback=gray!10, 
                  colframe=gray!40!black, 
                  leftrule=1pt,
                  rightrule=1pt,
                  toprule=1pt,
                  bottomrule=1pt,
                  arc=2mm,
                  title=\textbf{Prompt},
                  fonttitle=\bfseries,
                  top=1mm,
                  bottom=1mm,
                  toptitle=1mm,
                  bottomtitle=1mm]
\begin{center}
\itshape American cowboy with a scruffy appearance in a retrofuturistic style, inspired by the animations of Studio Ghibli.
\end{center}
\end{tcolorbox}

\begin{tabular}{ll}
$\bigcirc$ & \textbf{Image 1}  \\
$\bigcirc$ & \textbf{Image 2}  \\
$\bigcirc$ & \textbf{Tie} (both are equally aligned/misaligned)
\end{tabular}

\end{tcolorbox}
\caption{\textbf{User Study Template.} Instructions and question format used in our user study for evaluating image pairs.}
\label{fig:user-study-template}
\end{figure*}

\autoref{fig:user-study-template} shows the template used for our user study to evaluate image quality and prompt alignment across different sampling schedules. To ensure unbiased assessment, we employed a two-phase evaluation process for each image pair.

In our study methodology, participants first evaluated image quality (Question 1) without considering the prompt. Only after completing this quality assessment were they shown the prompt to evaluate text-image alignment (Question 2). This sequential approach ensured that prompt content did not influence quality judgments.

For each evaluation, we presented participants with pairs of images generated using different sampling schedules (Default, AYS, and OYS). The images were randomly shuffled and anonymized so participants had no knowledge of which sampling technique generated each image.

The results of this user study, presented in the main body of the paper, demonstrate a strong preference for images generated using our OYS method compared to both Default and AYS sampling schedules in terms of both image quality and prompt alignment.

\section{EDM2 Sampling Configurations}

We depict the parameters discovered by OYS and compare them with the default parameters in \autoref{tab:edm2_configs}. The clearest trend is in $\sigma_{\text{max}}$: OYS drives it down to 20 — the lower limit of our search range — for every model except EDM2-S, which settles just above the limit at 22.1. All are far below the default of 80. That the optimum saturates this bound suggests the preferred $\sigma_{\text{max}}$ may lie lower still, and that the default substantially overestimates the noise level at which sampling needs to begin. $\sigma_{\min}$ shows a complementary trend: every model saturates a limit of its $[0.0001, 0.01]$ search range — EDM2-XS the lower limit, and every other model the upper limit of 0.01, five times the default of 0.002. Together, these results suggest that OYS narrows the sampled noise range at both ends. Guidance splits by model scale: OYS raises it above the default for the smaller models (EDM2-XS through EDM2-M), but for the larger models (EDM2-L through EDM2-XXL) it drives guidance to 1.0 — again the edge of the search range — effectively disabling guidance altogether. The remaining parameters show no consistent trend across models, though we note that $\rho$ also saturates its upper limit of 10 for EDM2-L.

\begin{table}[!ht]
\centering
\small
\begin{tabular}{lccccc}
\toprule
\textbf{Model} & $\sigma_{\min}$ & $\sigma_{\max}$ & $\rho$ & \textbf{EMA length} & \textbf{Guidance} \\
\midrule
\multicolumn{6}{c}{\textbf{EDM2-XS}} \\
\midrule
Default & 0.002 & 80.0 & 7.0 & 0.045 & 1.40 \\
OYS & 0.0001 & 20.0 & 4.976 & 0.025 & 1.758 \\
\midrule
\multicolumn{6}{c}{\textbf{EDM2-S}} \\
\midrule
Default & 0.002 & 80.0 & 7.0 & 0.025 & 1.40 \\
OYS & 0.010 & 22.132 & 9.767 & 0.028 & 1.659 \\
\midrule
\multicolumn{6}{c}{\textbf{EDM2-M}} \\
\midrule
Default & 0.002 & 80.0 & 7.0 & 0.030 & 1.20 \\
OYS & 0.010 & 20.0 & 3.466 & 0.020 & 1.505 \\
\midrule
\multicolumn{6}{c}{\textbf{EDM2-L}} \\
\midrule
Default & 0.002 & 80.0 & 7.0 & 0.015 & 1.20 \\
OYS & 0.010 & 20.0 & 10.000 & 0.099 & 1.000 \\
\midrule
\multicolumn{6}{c}{\textbf{EDM2-XL}} \\
\midrule
Default & 0.002 & 80.0 & 7.0 & 0.020 & 1.20 \\
OYS & 0.010 & 20.0 & 4.098 & 0.099 & 1.000 \\
\midrule
\multicolumn{6}{c}{\textbf{EDM2-XXL}} \\
\midrule
Default & 0.002 & 80.0 & 7.0 & 0.015 & 1.20 \\
OYS & 0.010 & 20.0 & 4.901 & 0.077 & 1.000 \\
\bottomrule
\end{tabular}
\caption{\textbf{EDM2 Sampling Configurations.} Comparison of the default and OYS-optimized sampling configurations for each EDM2 model size.}
\label{tab:edm2_configs}
\end{table}

\section{Notation}

We summarize the notation used throughout the paper in \autoref{table:notation}.

\begin{table*}[t]
    \centering
    \begin{tabular}{@{}cl@{}}
        \toprule
        \textbf{Notation} & \textbf{Definition} \\
        \midrule
        \multicolumn{2}{c}{\textit{Problem setup}} \\
        \midrule
        $\mathcal{X}$ & Set of task inputs (e.g., text prompts) used for tuning \\
        $m$ & Sampling function of the fixed, pretrained diffusion model \\
        $e$ & Black-box evaluation metric \\
        $\mathbf{p}$ & Sampling configuration, $\mathbf{p} = \{p_1, \dots, p_M\}$; the parameters OYS optimizes \\
        $p_k$ & $k$-th element of $\mathbf{p}$; $p_k \in [0,1]$ for the schedule parameters \\
        $M$ & Number of parameters in $\mathbf{p}$ \\
        $\mathcal{P}$ & Search space of sampling configurations \\
        $\mathbf{p}^*$ & Optimal sampling configuration \\
        \midrule
        \multicolumn{2}{c}{\textit{Sampling schedule}} \\
        \midrule
        $K$ & Number of steps in the sampling schedule \\
        $s$ & Function mapping a configuration $\mathbf{p}$ to a sampling schedule \\
        $t_{\max}$ & Maximum timestep of the diffusion model \\
        $t_k$ & $k$-th timestep of the sampling schedule \\
        $q$ & Rounding to the nearest valid integer timestep, for discrete-time models \\
        $\sigma_{\min}, \sigma_{\max}, \rho$ & Parameters of the EDM parametric schedule \\
        \midrule
        \multicolumn{2}{c}{\textit{Bayesian optimization}} \\
        \midrule
        $f$ & Objective function, $f(\mathbf{p}) = e(m(\mathcal{X}; \mathbf{p}))$ \\
        $N$ & Number of observed evaluations \\
        $\mathbf{p}^{(j)}$ & $j$-th observed sampling configuration \\
        $y^{(j)}$ & Metric value observed for configuration $\mathbf{p}^{(j)}$ \\
        $\mathcal{D}$ & Observed configuration--performance pairs, $(\mathbf{p}^{(j)}, y^{(j)})_{j=1}^{N}$ \\
        \midrule
        \multicolumn{2}{c}{\textit{Human preference score}} \\
        \midrule
        $T$ & Text prompt \\
        $I$ & Generated image \\
        $\mathcal{E}_{T}$ & CLIP text encoder \\
        $\mathcal{E}_{I}$ & CLIP image encoder \\
        $\omega$ & Learned scaling parameter of the human preference score \\
        $\text{HPS}(T, I)$ & Human preference score of image $I$ for prompt $T$ \\
        $P(I_1 \succ I_2 \mid T)$ & Predicted preference win rate of image $I_1$ over $I_2$ given prompt $T$ \\
        \bottomrule
    \end{tabular}
    \caption{\textbf{Notation.} Summary of the notation used throughout the paper.}
    \label{table:notation}
\end{table*}

\end{document}